\documentclass[pdflatex]{article}

\usepackage{wacv}              

\usepackage{graphicx}
\usepackage{amsmath}
\usepackage{amssymb}
\usepackage{booktabs}
\usepackage{subcaption}
\usepackage{xcolor}
\usepackage{tcolorbox}
\usepackage{paracol}

\tcbset{
    mypromptbox/.style={
        colback=blue!5,        
        colframe=blue!75!black, 
        fonttitle=\bfseries,  
        title={Prompt},       
        sharp corners,        
        boxrule=1pt,          
        arc=4mm,              
        left=5mm,             
        right=5mm,            
        top=3mm,              
        bottom=3mm,           
    }
}
\newenvironment{cmtt}{\fontfamily{cmtt}\selectfont}{\par}
\usepackage{hyperref}

\usepackage[capitalize]{cleveref}
\crefname{section}{Sec.}{Secs.}
\Crefname{section}{Section}{Sections}
\Crefname{table}{Table}{Tables}
\crefname{table}{Tab.}{Tabs.}

\def\wacvPaperID{42} 
\def\confName{WACV}
\def\confYear{2025}

\begin{document}


\title{A Zero-Shot Learning Approach for Ephemeral Gully Detection from Remote Sensing using Vision Language Models}

\author{Seyed Mohamad Ali Tousi\textsuperscript{\textdagger\thanks{These authors contribute equally to the paper.}} , Ramy Farag\textsuperscript{\textdagger*}, Jacket Demby's\textsuperscript{\textdagger}, Gbenga Omotara\textsuperscript{\textdagger}, \\
John A. Lory\textsuperscript{\textdaggerdbl}, G. N. DeSouza\textsuperscript{\textdagger}\\
\small \textsuperscript{\textdagger}Vision Guided and Intelligent Robotics Laboratory (ViGIR), EECS Dept.\\
\small \textsuperscript{\textdaggerdbl}Division of Plant Sciences \\
\small University of Missouri
Columbia - MO - US\\
{\tt\small stousi, rmf3mc, udembys, goowdf, loryj, desouzag@missouri.edu}
}
\maketitle


\begin{abstract}
   Ephemeral gullies are a primary cause of soil erosion and their reliable, accurate, and early detection will facilitate significant improvements in the sustainability of global agricultural systems. In our view, prior research has not successfully addressed automated detection of ephemeral gullies from remotely sensed images, so for the first time, we present and evaluate three successful pipelines for ephemeral gully detection. Our pipelines utilize remotely sensed images, acquired from specific agricultural areas over a period of time. The pipelines were tested with various choices of Visual Language Models (VLMs), and they classified the images based on the presence of ephemeral gullies with accuracy higher than 70\% and a F1-score close to 80\% for positive gully detection. Additionally, we developed the \underline{first public dataset} for ephemeral gully detection, labeled by a team of soil- and plant-science experts. To evaluate the proposed pipelines, we employed a variety of zero-shot classification methods based on State-of-the-Art (SOTA) open-source Vision-Language Models (VLMs). In addition to that, we compare the same pipelines with a transfer learning approach. Extensive experiments were conducted to validate the detection pipelines and to analyze the impact of hyperparameter changes in their performance. The experimental results demonstrate that the proposed zero-shot classification pipelines are highly effective in detecting ephemeral gullies in a scenario where classification datasets are scarce.

\end{abstract}

\section{Introduction}

Soil erosion is a geomorphological land degradation process that can result in environmental harm, property damage, loss of livelihoods and services, and social and economic challenges. Ephemeral gully erosion reduces the sustainability of agricultural systems and causes significant sediment-related issues downstream \cite{soil}. Ephemeral gullies remain a prominent and hard-to-treat cause of  soil erosion \cite{gully_china}. A fast, reliable and cost-effective method to detect ephemeral gullies in the agricultural fields could play a crucial role in preventing the soil erosion and successful execution of soil conservation practices. A detection system using remote sensing images to distinguish between agricultural fields that show signs of ephemeral gully formation versus ephemeral gully-free agricultural fields will advance efforts of  soil researchers, farmers and responsible agencies to study, treat and/or prevent soil erosion. 


The emergence of Foundation Models, Vision Language Models (VLM), and Visual-Question-Answering (VQA) has great potential for tackling a wide range of computer vision challenges, particularly in remote sensing (\cite{guo2024remotesensingchatgptsolving, xu2024rsagentautomatingremotesensing, osco2023potentialvisualchatgptremote}). When classifying nuanced and subjective concepts, Foundation Models and VLMs tend to hallucinate. To address this problem, techniques such as Modeling Collaborator (MC) \cite{toubal2024modeling} have been introduced. One of the advantages of MC is that it does not require an extensive amount of labeled data which is especially attractive to us since labeled datasets for ephemeral gully detection are extremely scarce. The lack of sufficient datasets for detecting ephemeral gullies to train conventional computer vision models leads to the exploration of methods that do not rely heavily on extensive labeled data. Techniques such as zero-shot classification, and transfer learning are common alternatives. 

We assessed SOTA open-source zero-shot, transfer learning, and VLM-based classification methods in the task of classifying remote sensing images with signs of ephemeral gullies in agricultural fields. Additionally, we are providing the first dataset for detecting ephemeral gullies using remote sensing images, labeled by a group of soil and plant science experts. 

Our key contributions are as follows:

\begin{enumerate}
    \item We present and evaluate three classification pipelines for detecting ephemeral gullies in remote sensing images, exploring zero-shot, transfer learning, and Vision-Language Models (VLMs). To our knowledge, this is the first work to successfully address this under-explored task. Inspired by \cite{toubal2024modeling}, we further integrate large language models (LLMs) to enhance reasoning in the classification process, using VLMs for Visual Question Answering (VQA) and LLMs for reasoning, and compare their performance against existing zero-shot classification baselines.
    
    \item We conduct extensive experiments to investigate the choices of VLMs, types and number of questions, and aggregation methods for VQAs. Those experiments lead to a better understanding of the advantages and limitations of zero-shot classification methods such as \cite{toubal2024modeling}. 

    \item We make available the first public dataset for the detection of ephemeral gullies using remote sensing images. The dataset comprises high-resolution remote sensing RGB images of agricultural locations over different periods of time. The images are labeled by a team of soil and plant science experts to establish the ground truth classifications.

\end{enumerate}

\section{Background and Related Work}

In this section, we will explore the existing research articles in the literature on ephemeral gully detection (assuming its presence in the image), and ephemeral gully formation assessments using remote sensing imagery. We also explore the SOTA zero-shot classification methods and VLMs that form the foundation of this research.

\begin{figure*}
    \centering
    \includegraphics[width=0.9\linewidth]{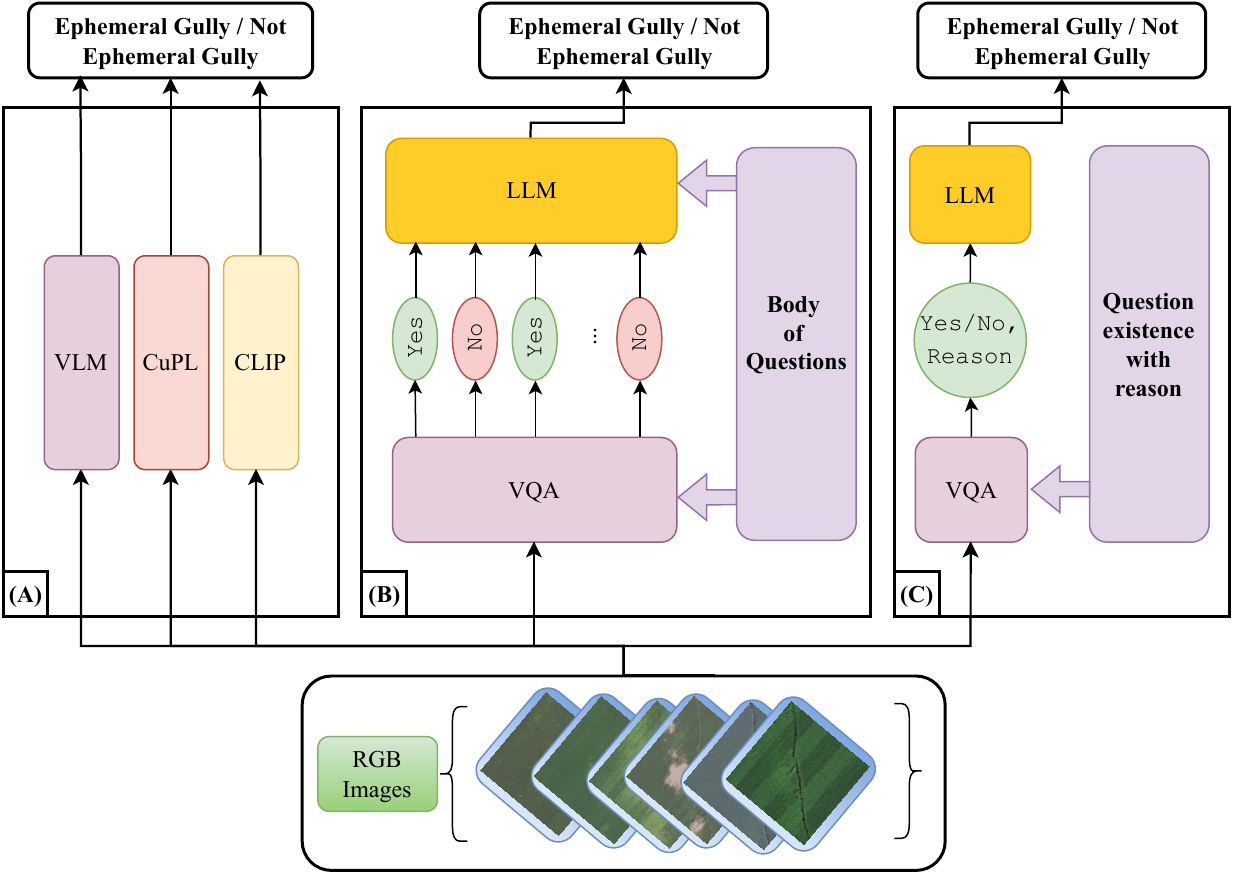}
    \caption{The proposed independent pipelines to detect ephemeral gullies. We are exploring three types of Pipelines;  \textbf{(A)}: Feeding the remote sensing RGB images directly to SOTA zero-shot classification methods such as CLIP \cite{radford2021learning}, CuPL \cite{pratt2023does}, and a variety of VLMs including Llama 3.2-Vision \cite{touvron2023llama}, Qwen \cite{bai2023qwen}, and Llava \cite{liu2023visual}. \textbf{(B)}: Inspired by \cite{toubal2024modeling}, we pass the RGB images through a VQA system, which responds to a series of Yes/No questions regarding visual attributes indicative of ephemeral gullies. An LLM then aggregates these responses to classify the images. \textbf{(C)}: Similar to (B), we propose asking the VLM a single descriptive question about the image, with an LLM interpreting the response to make a final classification.}
    \label{fig:pipeline}

    \vspace{-4mm}
\end{figure*}

\subsection{Ephemeral Gullies in Remote Sensing}

Remote detection of ephemeral gully through classification remains largely unexplored, though manual assessment and evaluation are well-documented in agricultural literature \cite{Identifying, w12020603, china1, china2}. However, the time commitment associated with manual  assessments limits their scalability for large-scale applications. Authors in \cite{reece2023using}  tested the effectiveness of using Google Earth temporal images to assess ephemeral gullies in U.S. agricultural fields, presenting a systematic method for identifying and locating these gullies within a certain error margin. Their approach relied heavily on manual labor and utilizes Google Earth images, which lack the image standardization needed for remote sensing in computer vision contexts.

Some researchers have attempted to automate parts of the process. There is extensive research on ephemeral gully risk assessment using topographic index models to identify areas of the field most susceptible to formation of ephemeral gullies \cite{sheshukov2018accuracy}. There are examples of computer vision  techniques applied to detect ephemeral gullies. For example,  \cite{ijgi6110371} employed conventional computer vision techniques, such as directional edge detection, to create a "semiautomatic" method for identifying ephemeral gullies. Manual pre-identification of areas containing gullies was necessary due to the faint visibility of these features. Similarly, \cite{LIU2022371cnn} applied deep learning methods like Convolutional Neural Networks (CNNs) \cite{krizhevsky2012imagenet} and U-Nets \cite{ronneberger2015uUnet} for delineating gullies in non-agricultural areas. Current model- and image-based methods require either post-identification verification of presence of a gully or manual pre-identification of gully-prone regions. To date, the core task of ephemeral gully classification continues to rely on human intervention.

\subsection{VLMs and Zero-Shot Classification}

Authors in \cite{liu2023visual, liu2024improved} introduced Visual Instruction Tuning through their LLaVA (Large Language and Vision Assistant) model, aimed at developing a general-purpose vision-language assistant. LLaVA integrates a CLIP-based visual encoder with a language model, specifically Vicuna \cite{vicuna2023, peng2023instruction}, fine-tuned using multimodal instruction-following data generated by GPT-4 \cite{achiam2023gpt}. This combination enables the model to effectively process and understand both visual and textual inputs, enhancing performance across diverse vision-language tasks.

Zero-shot classification allows deep learning models to classify instances without prior training on the same distribution of instances. Significant advancements in this field have been achieved through models like CLIP (Contrastive Language–Image Pre-training) \cite{radford2021learning}, which employs vast internet-sourced image-text pairs to align image and text embeddings in a unified multi-modal space. This alignment enables robust zero- and few-shot transfer performance for various vision tasks without fine-tuning. Extensions of CLIP \cite{pratt2023does, sun2024alpha, li2022grounded} continue to refine this alignment for improved downstream task performance. Building on this foundation,  \cite{saha2024improved} introduced AdaptCLIPZS, a framework that enhances zero-shot classification in fine-grained domains by incorporating detailed, category-specific descriptions generated by LLMs. These descriptions are integrated into context-rich prompts during VLM fine-tuning, consistently improving performance across benchmarks.

Modeling Collaborator (MC) \cite{toubal2024modeling} has demonstrated strong results in nuanced and subjective zero-shot classification tasks. Combining Visual-Question-Answering (VQA) with LLMs, MC avoids common large-model hallucinations in subjective tasks by using a series of Yes/No questions derived from the classification concept. These responses are aggregated by an LLM to produce labels that can be directly used or employed in training end classifiers. Unlike manual-label-heavy approaches such as Agile Modeling (AM) \cite{stretcu2023agile}, MC achieves comparable results with only 100 labeled samples for validation. This framework excels at classifying subjective visual concepts, outperforming traditional zero-shot and AM methods in both accuracy and efficiency.

The rest of the paper is organized as follows; In section \ref{methodology} we will present the proposed pipelines to detect ephemeral gully. The experimental setup, implementation details and experimental results will be presented in section \ref{experiments}. Then, in sections \ref{discussions} and \ref{conclusion} we will present a discussion about the experimental results and conclusion of the paper respectively.

\section{Proposed Method} \label{methodology}

Given a set of RGB images $I = \{i_1,i_2,...,i_M\}$ representing $M$ temporal images (tile) at one specific location, we create a dataset $\mathcal{D} = \{(I_n,y_n)| n\in \{1,2,...,N\}\}$ consisting of $N$ locations, in which each location has a true label $y_n$, determined by a group of soil and plant science experts looking at $I_n$. Figure \ref{fig:pipeline} shows the three classification pipelines proposed and evaluated in this work, which are presented in the next three sections. 

\subsection{SOTA Zero-Shot Classification Methods}

In the first classification pipeline that acts as the baseline for our application (pipeline \textbf{A} in Figure \ref{fig:pipeline}), we will use a variety of zero-shot classification methods, such as CLIP \cite{radford2021learning}, CuPL \cite{pratt2023does} and multiple VLMs like Llama 3.2 - Vision \cite{touvron2023llama}, Qwen \cite{bai2023qwen}, and Llava \cite{liu2023visual})

The remote sensing RGB images are submitted to those models to produce a label $\hat{y}_n$ for each location $I_n$. Equation \ref{eq:baseline} describes this pipeline.

\begin{equation}\label{eq:baseline}
    \hat{y}_n = VLM(I_n,p)
\end{equation}
Where $p$ is the input prompt to the VLM for detecting ephemeral gullies \footnote{CLIP \cite{radford2021learning} requires a positive and a negative prompt to produce the label for each $I_n$. Other VLMs can be prompted directly to produce the label.}.

\subsection{Visual Attributes + Reasoning}

Building on \cite{toubal2024modeling}, the second pipeline proposed and evaluated in this research is designed as an end-to-end process (see pipeline \textbf{B} in Figure \ref{fig:pipeline}). This pipeline consists of two key components: 1) \textit{Visual Attributes Understanding} and 2) \textit{Reasoning}. The initial step in this pipeline involves identifying distinctive visual attributes of ephemeral gullies to generate a \textit{body of questions} centered around these attributes. These questions help mitigate hallucinations when using baseline VLMs. The VLM processes the images alongside the generated questions, providing answers based on the image content. In the Reasoning component, these answers are aggregated by an LLM to determine the predicted class.

\subsubsection{Visual Attributes Understanding} \label{method:vis}

 The goal of the first component of pipeline \textbf{B}  is to:
\begin{enumerate}

    \item Identify specific visual attributes of a formed ephemeral gully. 
    
    \item Create a \textit{Body of Questions}, $Q = (q_1, q_2, ...,q_K)$ consisting of $K$ Yes/No questions based on the identified visual attributes. 

    \item Use the VLM to generate answers to $Q$ based on $I_n$.
    
\end{enumerate} 

In \cite{pratt2023does}, the authors proposed to use an LLM (e.g. GPT-4 \cite{achiam2023gpt}) to produce the specific visual attributes associated with each class. In this work, we use a combination of attributes generated by GPT-4 and visual aspects coming from plant and soil science experts. Once we obtain a set of visually distinguishable attributes for the ephemeral gully class, we build a body of questions in which we have one or more questions targeting each of the attributes.

Once $Q$ is obtained, answers are generated by a VLM:

\begin{equation}
    \boldsymbol{A}_n = VLM(I_n,Q)
\end{equation}

Where $\boldsymbol{A}_n$ is the set of answers for all of the Yes/No questions in $Q$, based on $I_n$ contents. Subsequently, we can proceed to do reasoning based on given answers to our questions.

\subsubsection{Reasoning} \label{method:reason}

In the reasoning component of pipeline \textbf{B}, an LLM model aggregates the answers of $Q$ and produce the final prediction:

\begin{equation}
    \hat{y}_n = LLM(\boldsymbol{A}_n,Q,p)
\end{equation}

Where $p$ is the aggregation prompt, and $\hat{y}_n$ is the predicted class for $I_n$.

\subsection{A Descriptive Question + Reasoning}

The last pipeline investigated in this research involves formulating a descriptive question to be submitted to the VLM for reasoning over the content of image location $I_n$ (refer to pipeline \textbf{C} in Figure \ref{fig:pipeline}). Subsequently, an LLM evaluates the provided reasoning to determine its validity and uses it to produce the final label $\hat{y}_n$.

\subsection{Transfer Learning Using Visual Attributes}

For comparison with the previous three proposed pipelines, we devised a non-zero-shot approach based on transfer learning. The rationale for this fourth approach was to contrast zero-shot and non-zero-shot methods with the intent of evaluating the performance of zero-shot methods. This approach uses the answers to questions $Q$ generated by the VLM as extracted features from the contents of image location $I_n$. These features are then used as inputs to train a simple Multi Layer Perceptron (MLP) on the development (Dev) set to produce the final label $\hat{y}_n$. The results of this approach are compared with the zero-shot methods discussed earlier, demonstrating that the zero-shot algorithms produce similar performance to this learning-based approach.

\section{Experimental Setup and Results} \label{experiments}

As mentioned earlier, the main contribution of our work are: the use of VLMs as zero-shot classification approaches for detecting ephemeral gullies; exploring different types of queries using VQA, and an extensive labeled dataset. In order to evaluate the efficacy of the proposed pipelines and the queries used for VQA, we must address the following questions:

\begin{enumerate}

    \item How well each pipeline performs the task of ephemeral gully detection?

    \item What are the impacts of specific choices of VLMs and LLMs in those pipelines?

    \item How the types and number of questions affect the performance of the pipelines?

    \item How does the performance of the proposed pipelines compare to a non-zero-shot approach based on transfer learning?

\end{enumerate}

To answer these questions, we conducted the following experiments to establish the validity and reliability of the classification pipeline.

\subsection{Evaluation Dataset}

We used USDA National Agriculture Imagery Program (NAIP) \cite{naip_usda} to gather 6 temporal high-resolution (1m) RGB images for agricultural areas over a period of 10 years. Google Earth Engine \cite{gorelick2017google} was used to access and download the dataset. Once the dataset was obtained, a team of 8 soil and plant science experts labeled each of the areas that showed a sign of ephemeral gully formation. Then, we divided the acquired dataset into two separate Dev and Test sets. 
Table \ref{tab:dataset} shows the statistics of the final evaluation dataset, after the labeling process and pruning the outliers. The dataset is available to download through this link: \url{http://vigir.missouri.edu/Research/GullyDetection/}. 


\begin{table}[]
\footnotesize
    \centering
    \begin{tabular}{@{}l|l|l@{}}
    \toprule
     \textbf{Parameter}            & \textbf{Dev Set} & \textbf{Test Set} \\
     \midrule
     Total Number of Locations         & 310              & 311              \\
     Number of EG-Positive Locations   & 177              & 177              \\
     Number of EG-Negative Locations   & 133              & 134              \\
     \midrule
     Number of Images per Location ($M$) & \multicolumn{2}{c}{6} \\
     Image Size                    & \multicolumn{2}{c}{128x128}          \\
     Image Spatial Resolution      & \multicolumn{2}{c}{1m}               \\
    \bottomrule
    \end{tabular}
    \caption{The Evaluation Dataset characteristics for Dev and Test sets.}
    \label{tab:dataset}
\end{table}


\subsection{Implementation Details}

This section provides the implementation details used to answer the research questions, as well as the experimental setups.

\subsubsection{Processing the RGB Images Simultaneously}

The primary implementation challenge to address was enabling VLMs to process six images as their inputs simultaneously. Since identifying ephemeral gullies relies on temporal information, processing each RGB image independently and aggregating the results is insufficient\footnote{This is the default way that CLIP \cite{radford2021learning} uses to process multiple images.}. Moreover, most of the open-source SOTA VLMs used in this study do not natively support multiple images at the time of writing. There are at least two potential solutions for this problem; The first solution involves encoding all six images into the embedding space, concatenating the embeddings, and then supplying the concatenated embedding vector as the conversation context when prompting the VLMs. However, this approach is constrained by the limited context length supported by VLMs—at most 4096 tokens at the time of writing—which makes it impractical to embed multiple separate RGB images.

The second solution is to create a collage of the images and pass it to the VLM as a single image. Our experiments prove this approach effective when working with various VLMs, including CLIP \cite{radford2021learning}, Llama 3.2-Vision \cite{touvron2023llama}, and Qwen \cite{bai2023qwen}.

\subsubsection{Body of Questions}

As mentioned in section \ref{method:vis}, we are using a combination of GPT-4 \cite{achiam2023gpt} and soil and plant science experts to produce the body of questions ($Q$). The final $Q$ is as follows:

\begin{tcolorbox}[mypromptbox, title=Body of Questions ($Q$)]
{\begin{cmtt}
\footnotesize
Given these six images of the exact same area and collected over a period of 10 years ...

\begin{enumerate}
    \item Do you see a low point in the terrain?
    \item Do narrow, winding paths or channels appear?
    \item Are there winding paths that become intermittent recurrent?
    \item Are there any linear depressions or ruts which appear more pronounced along natural drainage lines or slopes?
    \item Are there narrow and shallow channels which appear intermittently deeper or more indented into the soil?
    \item Are there areas where soil appears disturbed or vegetation is removed?
    
    \item Does a specific path lack vegetation, suggesting an evolving or emerging channel?
    \item Are there varying types and levels of coarseness in the texture of the soil?
    \item Are there clear starting and ending points of potential channels?
    \item Are there small rills or grooves indicating water flow?
    \item Is there a varying exposure of lighter or darker colored soil?
    \item Are there sediment accumulations forming?
    \item Are there signs of water activity, like soil clumps or crusting, that appear or intensify in specific areas?
    \item Are there any branching patterns that resemble temporary streams?
    \item Are there indications of nearby human activity, such as tillage or machinery tracks?
    \item Do you see any sign of water flow patterns across the field in multiple images?
    \item Do you see any edges in the images indicating removal of soil along the water pathway?



    \item Do you see any cuts in the soil associated with water flow across the field?
    \item Do you see any indication of human activity such as tillage that is not naturally happened in the field?
\end{enumerate}
\end{cmtt}}
\end{tcolorbox}

\vspace{-4mm}

\subsubsection{Experimental Setups and Model Choices}


We utilize LLava \cite{li2023m}, Llama 3.2-Vision \cite{touvron2023llama}, and Qwen \cite{bai2023qwen} as our VLMs. These models were selected primarily based on their performance on open benchmarks. At the time of writing, Llama 3.2 and Qwen rank among the top models across various vision and language benchmarks. LLava was included to facilitate a comparison between SOTA models and non-SOTA models.

We used Ollama \cite{ollama2024github} and Hugging Face \cite{wolf2019huggingface} as our frameworks to interact with the open-source VLMs and LLMs. Table \ref{tab:setup} shows the hyperparameters and model choices for the explored methods. 

To enable the VLMs to classify the presence of ephemeral gullies within a specific agricultural area, we  used two different prompts for pipelines \textbf{A}, and \textbf{C}: 

\vspace{-2mm}

\begin{tcolorbox}[mypromptbox, title=Prompt for Baseline Results (Pipeline \textbf{A})]
{\begin{cmtt}
\footnotesize
Given this collage of six images of the exact same area and collected over a period of 10 years. Are there any ephemeral gully appearances by looking at all of them together? Reason and conclude with only yes or no.
\end{cmtt}} 
\end{tcolorbox}

\vspace{-4mm}

\begin{tcolorbox}[mypromptbox, title=Prompt for Single Question Reasoning (Pipeline \textbf{C})]
{\begin{cmtt}

\footnotesize
Given this collage of six images of the exact same area and collected over a period of 10 years. Are there any ephemeral gully appearances by looking at all of them together?. Provide the reasons for your answer.
\end{cmtt}} 
\end{tcolorbox}

\begin{table}[]
\footnotesize
    \centering
    \begin{tabular}{@{}l|l@{}}
    \toprule
     \textbf{Parameter}            & \textbf{Value}  \\
     \midrule
    Number of Temporal Images ($M$) & 6 \\
    Number of Questions ($Q$ size) & 19 \\
    \midrule
     & Llama 3.2 - Vision \cite{touvron2023llama} \\
    VLM Choices & Qwen \cite{bai2023qwen}\\
     & Llava \cite{liu2023visual}\\
     \midrule
    LLM Choices & Llama 3.2 \cite{touvron2023llama}\\
    & Qwen \cite{bai2023qwen}\\
    \bottomrule
    \end{tabular}
    \caption{The experimental setup and model choices for exploring classification pipelines.}
    \label{tab:setup}
\end{table}

\subsection{Experimental Results}

To evaluate the performance of the classification pipelines and to address research questions \textbf{(1)} and \textbf{(2)}, we conducted experiments measuring the accuracy of each pipeline on the test set and analyzed the impact of different model choices for VLMs and LLMs. Additionally, to address question \textbf{(4)}, we trained a simple MLP on the Dev set to aggregate answers from the VQA step, with results presented in Table \ref{tab:vlm-perform}. Given the dataset's imbalance in positive and negative images, relying solely on the F1 score could be misleading; therefore, we included the Macro F1 score in Table \ref{tab:vlm-perform} to reflect a class-averaged evaluation.

\begin{figure*}[t!]
    \centering
    \begin{subfigure}[t]{0.48\columnwidth}
        \centering
        \includegraphics[width=\textwidth,trim={1cm 0cm 1.5cm 0cm}]{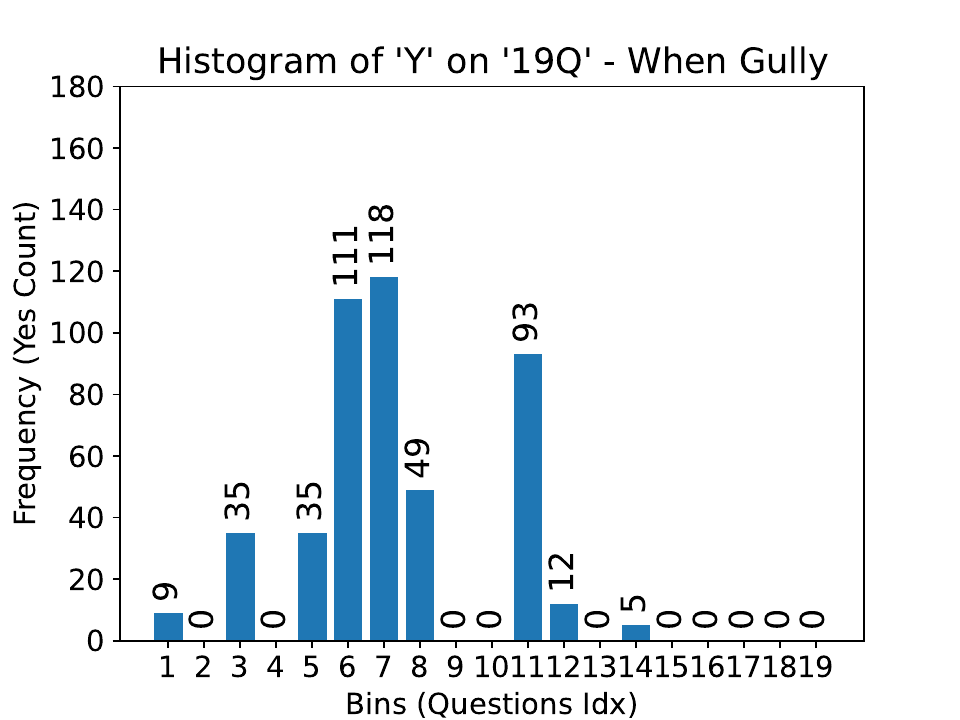}
        \caption{Yes - Gully (Valid Set)}
    \end{subfigure}%
    \hspace{2mm}
    \begin{subfigure}[t]{0.48\columnwidth}
        \centering
        \includegraphics[width=\textwidth,trim={1cm 0cm 1.5cm 0cm}]{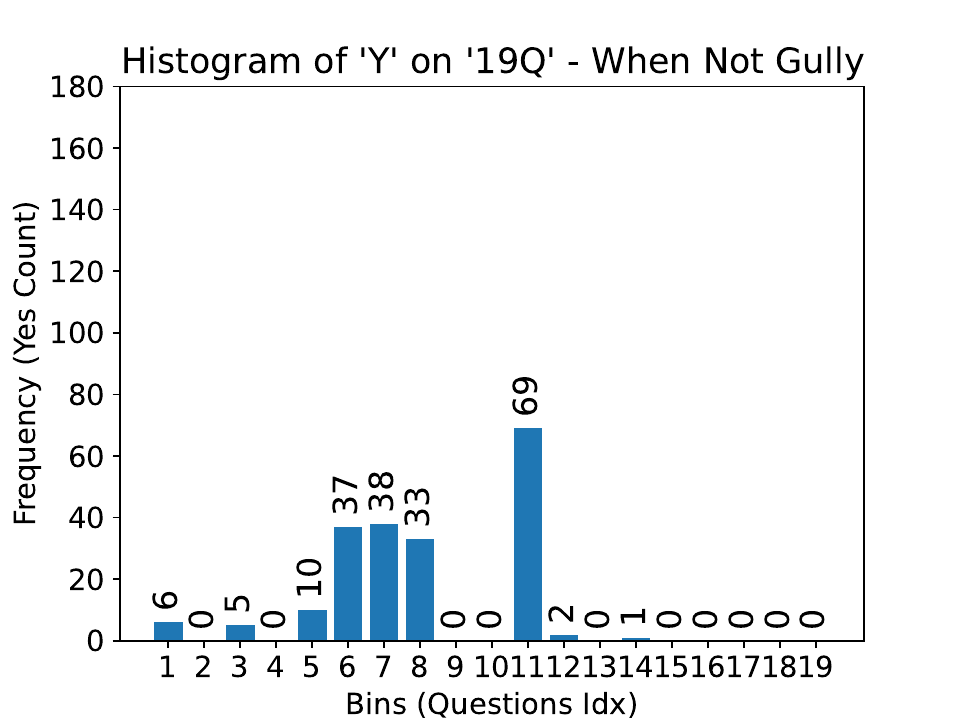}
        \caption{Yes - Not Gully (Valid Set)}
    \end{subfigure}
    \hspace{2mm}
    \begin{subfigure}[t]{0.48\columnwidth}
        \centering
        \includegraphics[width=\textwidth,trim={1cm 0cm 1.5cm 0cm}]{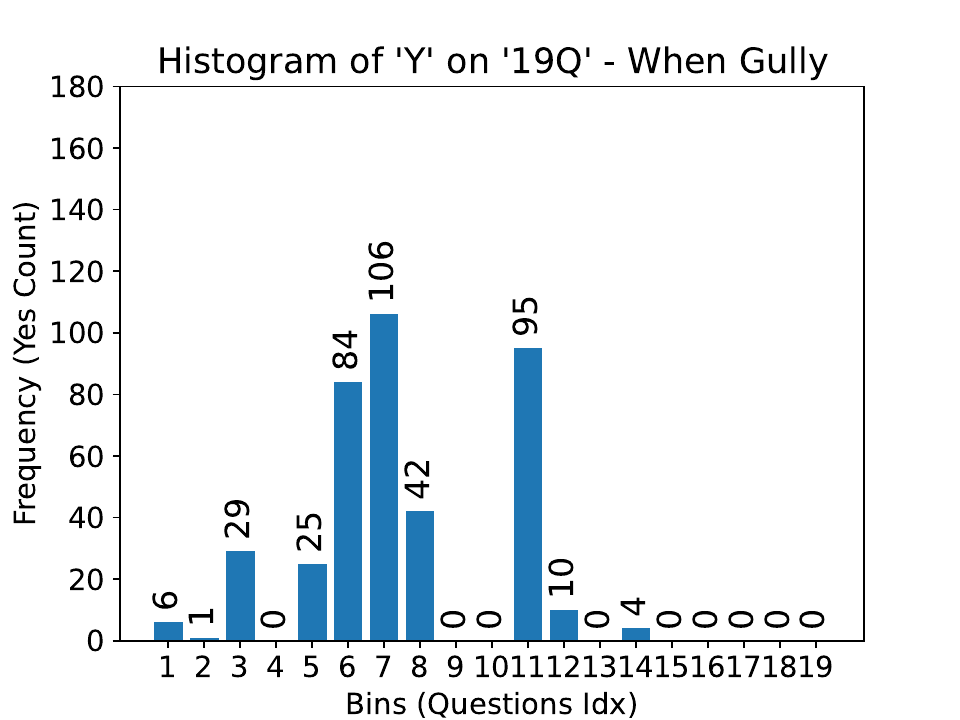}
        \caption{Yes - Gully (Test Set)}
    \end{subfigure}%
    \hspace{2mm}
    \begin{subfigure}[t]{0.48\columnwidth}
        \centering
        \includegraphics[width=\textwidth,trim={1cm 0cm 1.5cm 0cm}]{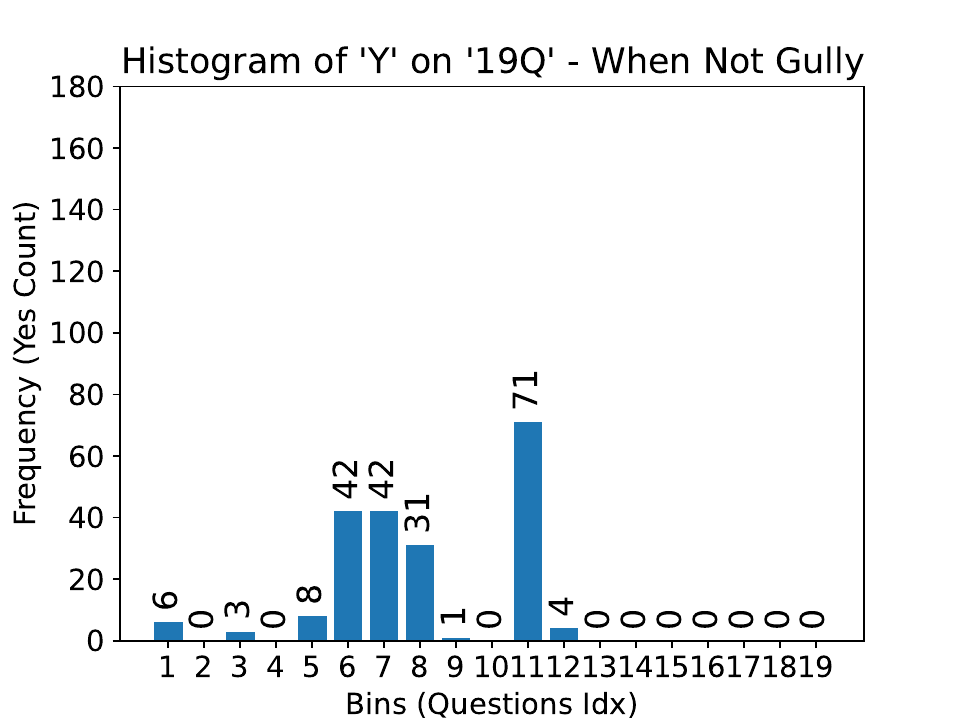}
        \caption{Yes - Not Gully (Test Set)}
    \end{subfigure}
    \caption{Analysis of responses across our body of 19 questions obtained with Llama3.2-90b VLM on the test set.}
    \label{fig:analyzing_questions}
\end{figure*}

To address research question \textbf{(3)}, we performed two additional experiments. First, an expert ranked questions in $Q$ by their relevance for distinguishing ephemeral gullies, creating subsets of 3, 6, 9, 12, 15, and 18 questions, with their indices listed in Table \ref{tab:parition-Q}. Second, a hyperparameter optimization framework (Optuna \cite{optuna_2019}) was used to identify the optimal set of questions for improving pipeline \textbf{B}. Performance results for these experiments are summarized in Table \ref{tab:vlm-Q}. To further analyze the decision-making process, a histogram in Figure \ref{fig:analyzing_questions} illustrates the frequency of "Yes" responses for each question across positive and negative test and Dev sets.

\begin{table*}[!t]
\footnotesize
  \centering
  \begin{tabular}{@{}p{2mm}lll|p{3mm}p{3mm}p{3mm}p{3mm}ll|llll@{}}
    \toprule
    \textbf{P\#} & \textbf{VLM} & \textbf{LLM} & \textbf{Experiment} & TP & FP & FN & TN & Prec.  & Rec.  & Acc.  & F1 (G) & F1 (NG)  & Macro F1\\
    \midrule 
    \midrule 
    \multicolumn{14}{c}{Input = Collage of 6 Temporal Aerial Images} \\
    \midrule
    \textbf{(A)} & CuPL & - & CuPL (GPT4-Prompts) & 9 & 4 & 168 & 130 & 0.692 & 0.051 & 0.447 & 0.095 & 0.602 & 0.348 \\

    \textbf{(A)} & CLIP & - & CLIP (Pos/Neg) & 134 & 90 & 43 & 44 & 0.598 & 0.757 & 0.572 & 0.668 & 0.398 & 0.533 \\

    \textbf{(A)} & Llava-Llama3-8b & - & VQA (1Y/N - Q\&A) & 22 & 20 & 155 & 114 & 0.524 & 0.124 & 0.437 & 0.201 & \underline{0.566} & 0.383 \\

    \textbf{(A)} & Llama3.2-90b & - & VQA (1Y/N - Q\&A) & 146 & 103 & 31 & 31 & 0.586 & 0.825 & 0.569 & 0.685 & 0.316 & 0.501 \\

    \textbf{(A)} & Qwen2-VL-72B & - & VQA (1Y/N - Q\&A) & 167 & 79 & 10 & 55 & 0.679 & 0.944 & \underline{\textbf{0.714}} & \underline{\textbf{0.790}} & 0.553 & \underline{\textbf{0.671}} \\

    \midrule

    \textbf{(B)} & Llava-Llama3-8b & Llama3.2 & VQA(15Y/N - Q\&A) & 177 & 134 & 0 & 0 & 0.569 & 1.000 & 0.569 & 0.725 & 0.000 & 0.363 \\

    \textbf{(B)} & Qwen2-VL-72B & Llama3.2 & VQA(15Y/N - Q\&A) & 173 & 103 & 4 & 31 & 0.627 & 0.977 & \underline{0.656} & \underline{0.764} & 0.367 & 0.565 \\

    \textbf{(B)} & Llama3.2-Vision & Qwen2 & VQA(15Y/N - Q\&A) & 25 & 7 & 152 & 127 & 0.781 & 0.141 & 0.489 & 0.239 & 0.615 & 0.427 \\
    
    \textbf{(B)} & Qwen2-VL-72B & Llama3.2 & VQA(15Y/N - Q\&A) & 177 & 134 & 0 & 0 & 0.569 & 1.000 & 0.569 & 0.725 & 0.000 & 0.363\\

    \textbf{(B)} & Llama3.2-90b & Llama3.2 & VQA(15Y/N - Q\&A) & 110 & 44 & 67 & 90 & 0.714 & 0.621 & 0.643 & 0.665 & \underline{\textbf{0.619}} & \underline{0.642} \\

    \midrule


   

    \textbf{(C)} & Llava-Llama3-8b & Llama3.2 & VQA (1Q + Reason) & 177 & 134 & 0 & 0 & 0.569 & 1.000 & 0.569 & 0.725 & 0.000 & 0.363 \\

    
    \textbf{(C)} & Llama3.2-90b & Llama3.2 & VQA (1Q + Reason) & 130 & 101 & 47 & 33 & 0.563 & 0.734 & 0.524 & 0.637 & \underline{0.308} & \underline{0.473} \\
    
    \midrule
    \textbf{(TL)} & Llama3.2-90b & - & VQA(15Y/N) + MLP & 117 & 50 & 60 & 84 & 0.701 & 0.661 & 0.646 & 0.680 & 0.604 & 0.642 \\

     \bottomrule
  \end{tabular}
  \caption{Performance of VLMs on the Ephemeral Gully classification dataset. P\# column indicates the pipelines as designated in Figure \ref{fig:pipeline}. In each pipeline, the best metric scores are underlined. The best metric scores across all the pipelines are in bold. (TL) stands for the transfer learning method which is training an MLP on the Dev set to aggregate the VQA results.}
  \label{tab:vlm-perform}
\end{table*}

\begin{table*}[!t]
\footnotesize
  \centering
  \begin{tabular}{@{}ll@{}}
    \toprule
    \textbf{Experiment} & \textbf{Question Indices} \\
    \midrule 
    \midrule 
    VQA(3Y/N - Q\&A + LLM)  & 2, 3, 14\\
    VQA(6Y/N - Q\&A + LLM)  & 2, 3, 14, 5, 4, 6\\
    VQA(9Y/N - Q\&A + LLM)  & 2, 3, 14, 5, 4, 6, 9, 11, 13 \\
    VQA(12Y/N - Q\&A + LLM) & 2, 3, 14, 5, 4, 6, 9, 11, 13, 8, 10, 7 \\
    VQA(15Y/N - Q\&A + LLM) & 2, 3, 14, 5, 4, 6, 9, 11, 13, 8, 10, 7, 1, 12, 15 \\
    VQA(18Y/N - Q\&A + LLM) & 2, 3, 14, 5, 4, 6, 9, 11, 13, 8, 10, 7, 1, 12, 15, 16, 17, 18 \\
    \bottomrule
  \end{tabular}
  \caption{Partition of the 19 Questions per index for the experiments in Table \ref{tab:vlm-Q}.}
  \label{tab:parition-Q}
\end{table*}


\begin{table*}[!t]
\footnotesize
  \centering
  \begin{tabular}{@{}llllllll|llll@{}}
    \toprule
    \textbf{Model} & Experiment & TP & FP & FN & TN & Prec.  & Rec.  & Acc.  & F1 (G) & F1 (NG)  & Macro F1\\
    \midrule 
    \midrule 
    \multicolumn{12}{c}{Collage of 6 Temporal Aerial Images} \\
    \midrule
    \textbf{Ours (Llama3.2-90b)} & VQA(3Y/N - Q\&A + LLM) & 29 & 3 & 148 & 131 & 0.906 & 0.164 & 0.514 & 0.278 & 0.634 & 0.456 \\
    \textbf{Ours (Llama3.2-90b)} & VQA(6Y/N - Q\&A + LLM) & 40 & 9 & 137 & 125 & 0.816 & 0.226 & 0.531 & 0.354 & 0.631 & 0.493 \\
    \textbf{Ours (Llama3.2-90b)} & VQA(9Y/N - Q\&A + LLM) & 37 & 9 & 140 & 125 & 0.804 & 0.209 & 0.521 & 0.332 & 0.627 & 0.479 \\
    \textbf{Ours (Llama3.2-90b)} & VQA(12Y/N - Q\&A + LLM) & 111 & 44 & 66 & 90 & 0.716 & 0.627 & 0.646 & 0.669 & 0.621 & 0.645 \\
    \textbf{Ours (Llama3.2-90b)} & VQA(15Y/N - Q\&A + LLM) & 110 & 44 & 67 & 90 & 0.714 & 0.621 & 0.643 & 0.665 & 0.619 & 0.642 \\
    \textbf{Ours (Llama3.2-90b)} & VQA(18Y/N - Q\&A + LLM) & 37 & 10 & 140 & 124 & 0.787 & 0.209 & 0.518 & 0.330 & 0.623 & 0.477 \\

\midrule
    \textbf{Optuna (Llama3.2-90b)} & VQA(4Y/N - Q\&A + LLM) & 110 & 44 & 67 & 90 & 0.714 & 0.621 & 0.643 & 0.665 & 0.619 & 0.642 \\


    $\Rightarrow$ \textbf{4Y/N: (3,6,7,12)} &  & & & && &  &  &  &  & \\

     \bottomrule
  \end{tabular}
  \caption{Performance of VLMs on the Ephemeral Gully classification dataset when varying the number of questions.}
  \label{tab:vlm-Q}
\end{table*}

\begin{figure}[t!]
    \centering
    \begin{subfigure}[t]{0.45\columnwidth}
        \centering
        \includegraphics[width=\textwidth,trim={1cm 0cm 1.5cm 0cm}]{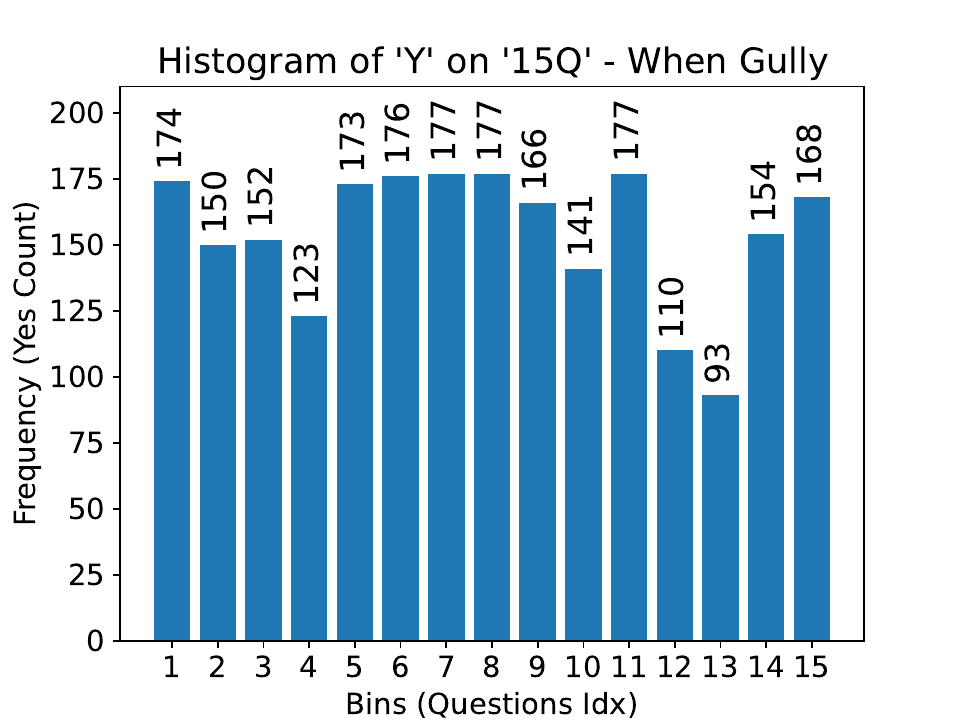}
        \caption{Yes - Gully (Test Set)}
    \end{subfigure}%
    \hspace{2mm}
    \begin{subfigure}[t]{0.45\columnwidth}
        \centering
        \includegraphics[width=\textwidth,trim={1cm 0cm 1.5cm 0cm}]{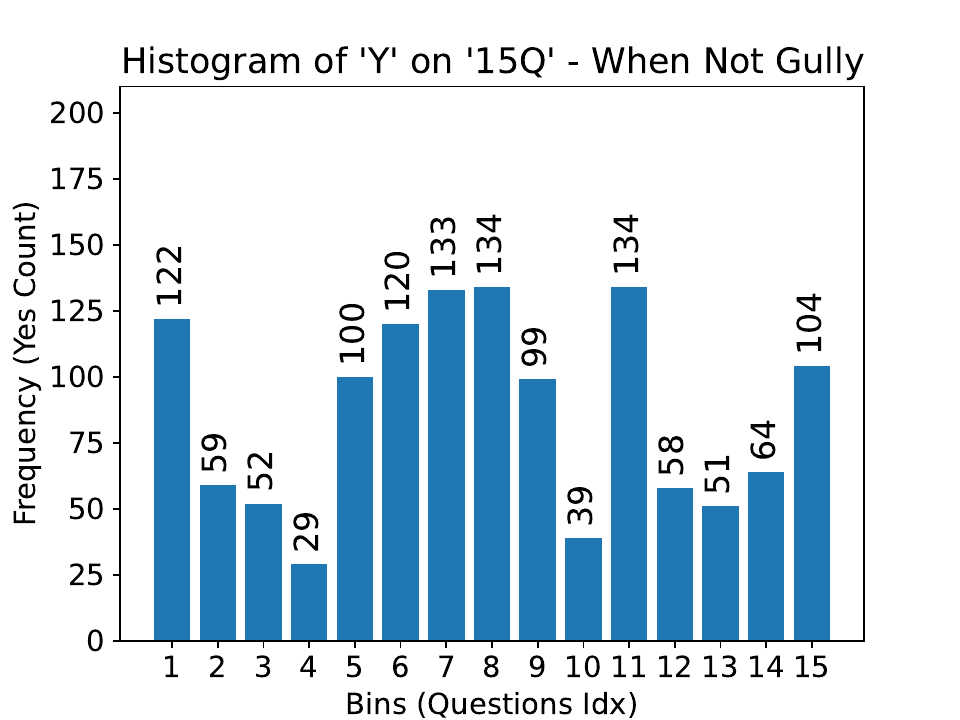}
        \caption{Yes - Not Gully (Test Set)}
    \end{subfigure}
    \caption{Analysis of responses across our body of 15 questions obtained with Qwen2-VL-72b VLM on the test set.}
    \label{fig:analyzing_questions_qwen}
\end{figure}

\section{Discussion} \label{discussions}

\subsection{Performance of The Classification Pipelines}

Both CLIP  and CuPL  presented difficulty detecting ephemeral gullies (Table \ref{tab:vlm-perform}). This  was expected given the limited representation of remote sensing images with delineated ephemeral gullies in their pre-training datasets. CuPL, which builds on CLIP, inherited the same challenges in distinguishing ephemeral gullies. Similarly, Llava  and Llama 3.2-Vision performed poorly in pipeline \textbf{A} due to biases in their classification behavior. Llava classified most areas as negatives, yielding a low F1 score for the EG-positive class, while Llama 3.2-Vision labeled most locations as EG-positive, reducing the F1 score for the EG-negative class. Both models exhibited low Macro F1 scores, highlighting the importance of pre-training datasets in zero-shot classification methods, as these datasets lacked sufficient information about ephemeral gullies' visual characteristics.

In contrast, Qwen2-VL demonstrated strong performance in pipeline \textbf{A}, suggesting that its pre-training dataset contained richer information about ephemeral gullies. We have also achieved promising results in pipeline \textbf{B}, particularly when using Llama 3.2-Vision as the VLM and Llama 3.2 as the LLM, performing comparably to Qwen2 in pipeline \textbf{A}. However, substituting either the VLM or LLM with Qwen2 reduced performance, as discussed in Section \ref{dis:QL}. Pipeline \textbf{C} did not surpass the performance of pipeline \textbf{A}, indicating that prompting VLMs to reason about their decisions may exacerbate hallucination rather than mitigate it. Additionally, a transfer learning-based method failed to outperform zero-shot classifiers in pipeline \textbf{A} (with Qwen2-VL) and pipeline \textbf{B} (with Llama 3.2 - Vision), suggesting that zero-shot methods are particularly effective for detecting ephemeral gullies in data-limited scenarios.

\subsection{Importance of Questions in Pipeline (B)}

Figure \ref{fig:analyzing_questions} illustrates the frequency of "Yes" responses for each question across EG-positive and EG-negative locations. The histogram reveals that not all questions significantly impact the decision-making process, which aligned with our expectations. This is because the questions, generated using expert input and GPT-4 knowledge, do not always correspond to visual features that VLMs can reliably interpret. A subset of the questions likely contributes more meaningfully to the results, leading to a series of experiments with different subsets of questions (Table \ref{tab:vlm-Q}). Performance improved as the number of questions increased, introducing critical visual attributes—up to a certain point. Beyond 12 questions, performance began to plateau, with a slight drop observed from 12 to 15 questions, and a significant decline from 15 to 18 questions.

This behavior can be attributed to two factors. First, as the number of questions grows, redundancy becomes more likely, leading to inconsistent answers that confuse the LLM during aggregation. Second, the increased number of questions raises the potential for conflicts among them, complicating the decision-making process. To address this, Optuna was used to identify the most influential questions. Remarkably, using just the four key questions identified by Optuna produced the same results as using 15 questions, demonstrating the effectiveness of optimizing the question set for better performance (Table \ref{tab:vlm-Q}). This is a clear advantage of our pipeline over other VQA-based classification approaches like MC \cite{toubal2024modeling}.

\subsection{Qwen2-VL vs. Llama3.2-Vision}\label{dis:QL}

Both Qwen2-VL and Llama 3.2-Vision showed strong performance in ephemeral gully detection tasks, but their strengths varied depending on the part of the pipeline they were used in (Table \ref{tab:vlm-perform}). Qwen2-VL excelled in pipeline \textbf{A}, demonstrating a robust ability to directly detect ephemeral gullies. Its superior visual understanding aligned with benchmarks \cite{2023opencompass}, which consistently rank Qwen2-VL above other open-source VLMs in visual comprehension. However, its language abilities lagged behind those of Llama 3.2, which makes the latter more effective in tasks requiring nuanced language reasoning.

Llama 3.2's linguistic proficiency enhanced its performance as an aggregator LLM, as shown in Table \ref{tab:vlm-perform}. Its superior language comprehension enabled it to generate better answers to the body of questions, making it a more effective VQA system. Conversely, Qwen2-VL's answer histograms (Figure \ref{fig:analyzing_questions_qwen}) revealed a tendency to respond "Yes" to most questions, especially the key questions identified by Optuna \cite{optuna_2019}, regardless of context. This overgeneralization limited its ability to distinguish between EG-positive and EG-negative tiles, often leading to the incorrect labeling of all tiles as positive.

\subsection{Practical Implications and Future Works}

The proposed pipelines demonstrated reliable and robust results; however, challenges remain that need to be addressed for their successful deployment in agricultural applications and real-world scenarios. In practical contexts, false negative classifications pose a greater concern than false positives, as missing ephemeral gullies in agricultural fields can lead to significant issues. To enhance the effectiveness of the proposed pipelines, improving their sensitivity (F1 (G)) to the presence of ephemeral gullies is crucial. Future research could explore the fine-tuning of Vision-Language Models (VLMs) to provide better and more robust vision and language understanding in this specific context, thereby improving the pipeline's overall performance and reliability.

\section{Conclusion} \label{conclusion}

This study introduced the first stand-alone pipeline to detect ephemeral gullies and an evaluation dataset labeled by experts. Our results demonstrate that VLM-based zero-shot classification methods, particularly those using Qwen2-VL and Llama 3.2 - Vision, were effective in detecting ephemeral gullies. While Qwen2-VL excelled in visual understanding, Llama 3.2 - Vision's language abilities made it a stronger aggregator in question-based methods. We observed that the number and quality of questions significantly impacted performance, with redundancy and conflicts in larger question sets leading to diminished results. Transfer learning approaches also provided similar, and in some cases lower performance compared to zero-shot methods that suggests the reliability and versatility of the proposed detection pipeline. 

\section{Acknowledgment}
Computational resources for this research have been supported by the NSF National Research Platform, as part of GP-ENGINE (award OAC \#2322218)

{\small
\bibliographystyle{ieeetr}
\bibliography{egbib}

}

\end{document}